\documentclass[runningheads]{llncs}
\usepackage{amsmath}
\usepackage{amsfonts}
\usepackage{dsfont}
\usepackage{graphicx}
\usepackage{tikz}
\usepackage{multirow}
\usepackage{array}
\usepackage{tabularx,booktabs}
\usepackage{subcaption}
\usepackage{dirtytalk}
\usepackage{bm}
\usepackage{dsfont}
\usepackage{float}
\usetikzlibrary{arrows.meta}
\usepackage{color}
\usepackage{hyperref}

\begin{document}

\title{Identifying Critical Tokens for\\Accurate Predictions in\\Transformer-based Medical Imaging Models}

\titlerunning{Identifying Critical Tokens for Accurate Predictions}

\author{Solha Kang\inst{1} \and
Joris Vankerschaver\inst{1,2} \and
Utku Ozbulak\inst{1,3}}

\authorrunning{Kang et al.}

\institute{
Center for Biosystems and Biotech Data Science,\\ Ghent University Global Campus, Republic of Korea \and
Department of Applied Mathematics, Computer Science and Statistics,\\ Ghent University, Belgium \and
Department of Electronics and Information Systems,\\ Ghent University, Belgium\\
\email{utku.ozbulak@ghent.ac.kr}
}

\maketitle

\begin{abstract}
With the advancements in self-supervised learning (SSL), transformer-based computer vision models have recently demonstrated superior results compared to convolutional neural networks (CNNs) and are poised to dominate the field of artificial intelligence (AI)-based medical imaging in the upcoming years. Nevertheless, similar to CNNs, unveiling the decision-making process of transformer-based models remains a challenge. In this work, we take a step towards demystifying the decision-making process of transformer-based medical imaging models and propose \say{Token Insight}, a novel method that identifies the critical tokens that contribute to the prediction made by the model. Our method relies on the principled approach of token discarding native to transformer-based models,  requires no additional module, and can be applied to any transformer model. Using the proposed approach, we quantify the importance of each token based on its contribution to the prediction and enable a more nuanced understanding of the model's decisions. Our experimental results which are showcased on the problem of colonic polyp identification using both supervised and self-supervised pretrained vision transformers indicate that Token Insight contributes to a more transparent and interpretable transformer-based medical imaging model, fostering trust and facilitating broader adoption in clinical settings.
\end{abstract}

\section{Introduction}

While pretrained CNNs~\cite{resnet,VGG} have been the go-to models for solving complex medical imaging problems in the past decade, the emergence of transformer-based architectures such as the vision transformer (ViT)~\cite{vit} and the self-attention mechanism~\cite{attention} originating from natural language processing (NLP) has significantly influenced the computer vision community, marking a shift towards replacing CNNs as the preferred models. The emergence of foundational models as well as other advances in SSL frameworks such as DINO~\cite{dino}, MAE~\cite{mae}, MoCo~\cite{moco_v3}, and many more~\cite{ozbulak2023know} has further sped up the adoption rate of transformer-based models. Nowadays, such models are employed in medical imaging to solve a variety of medical-imaging problems, with recent studies suggesting that ViTs generally outperform CNNs \cite{matsoukas2021time,shamshad2023transformers,singhal2023towards}.

Unfortunately, similar to other types of neural networks, transformer-based networks also struggle in in reasoning capabilities~\cite{attention_not_interpretable,is_att_interpretable}. That is, it is not straightforward to decipher the reason for predictions that have been made by transformer networks. To address the issue of lack of transparency for transformers, a number of novel interpretability methods for ViTs have been proposed, with some of these efforts making use of techniques borrowed from research on CNNs~\cite{chefer2021transformer}, while others focus on new areas native to transformers, such as the usage of attention maps and token-based methods~\cite{rigotti2021attention,attention_vs_saliency,not_all_tokens}. These interpretability methods often produce what are called interpretability maps, with the goal of highlighting the salient regions of the image. In the context of medical imaging, these regions often correspond to areas where the disease occurs (for example, highlighting a tumor). As a result, the performance of such methods is often evaluated based on their ability to highlight medically-relevant parts that are annotated by experts. However, recent research on spurious correlations suggests that the salient parts of the medical images identified by models might not be the same as the ones identified by experts~\cite{sun2023right}. Indeed, certain undesired signals may lead to what is called shortcut learning and result in models with undesired behaviors that perform the task at hand via signals that are not intended~\cite{geirhos2020shortcut}.

\begin{figure}[t!]
    \centering
    \includegraphics[width=0.75\textwidth]{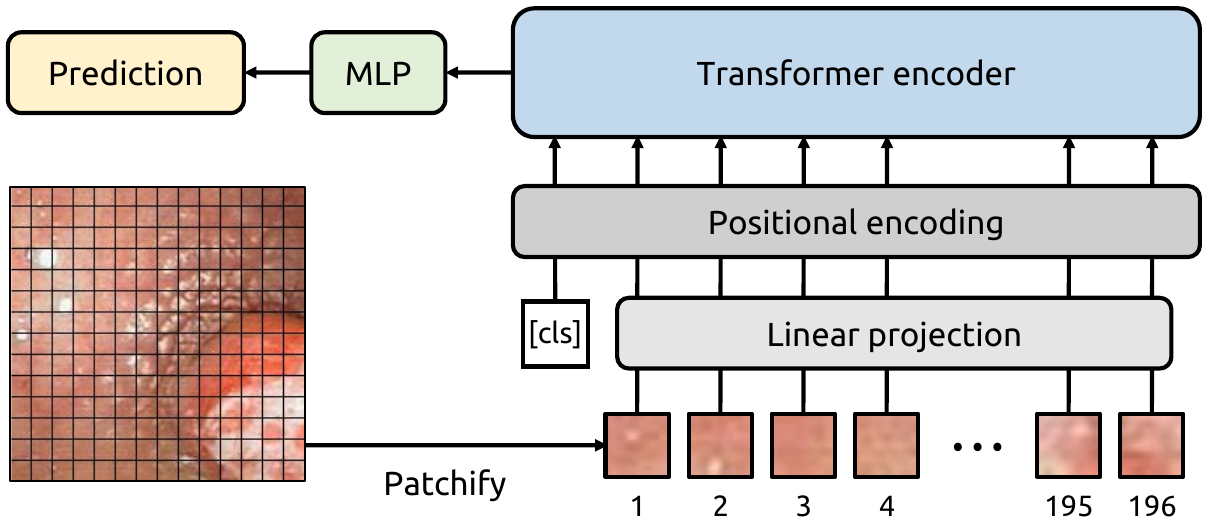}
    \caption{An overview of the ViT architecture and the tokenization of image patches are illustrated.}
    \label{fig:transformer_fig}
\end{figure}

To address these challenges with predictive reasoning, how can we identify the parts of medical images that lead to the predictions the model has made, regardless of undesired model behaviors such as shortcut learning? This work is centered around answering this question. Our method, called \say{Token Insight}, identifies the critical tokens that contribute to the predictions made by a ViT. This method relies on the principled approach of token discarding in ViTs, which has garnered significant interest for several reasons in recent years, for example to speed up training~\cite{token_merging1,token_remove2}, to create robust models~\cite{token_merging2}, as well as to provide interpretability~\cite{which_tokens}. In contrast to most interpretability methods that use a variety of metrics as a proxy for predictive reasoning (such as the overlap with salient regions), our method directly uses the reduction in prediction confidence as a means to identify critical tokens. As a result, Token Insight provides a more transparent and intuitive understanding of the model's decision-making process, leading to the identification of undesired model behaviors such as shortcut learning, offering a more reliable means to pinpoint the specific aspects of tokens influencing the model's predictions.

\section{Methodology}

In this section, we provide a description of the dataset used, outline the models utilized, and finally describe the proposed method in detail.

\begin{table}[t]
\centering
\caption{Details of the CP-CHILD-A and CP-CHILD-B datasets.}\label{tbl:dataset}
\begin{tabular}{cccccc}
\cmidrule[0.75pt]{2-6}
~ & \multicolumn{2}{c}{Training} & \multicolumn{2}{c}{Testing} & \multirow{3}{*}{Device} \\
\cmidrule[0.75pt]{1-5}
Dataset & \phantom{--}Non-polyp\phantom{--} & \phantom{--}Polyp\phantom{--} & \phantom{--}Non-polyp\phantom{--} & \phantom{--}Polyp & ~ \\
\midrule
CP-CHILD-A & 6,200 & 800 & 800 & 200 & Olympus PCF-H290DI\\
CP-CHILD-B & 800 & 300 & 300 & 100 & FUJIFILM EC-530wm \\
Combined & 7,000 & 1,100 & 1,100 & 300 & Both\\
\bottomrule
\end{tabular}
\end{table}

\subsection{Dataset}
\label{sec:dataset}

For straightforward and easily understandable experiments, we utilize the combination of recently proposed CP-CHILD-A and CP-CHILD-B datasets, which involves the detection of colonic polyps~\cite{wang2020improved}. Both datasets combined comprises of 9,500 colonoscopy RGB images obtained from 1,600 patients using the Olympus PCF-H290DI and FUJIFILM EC-530wm. The task involves identifying colonic polyps, which is challenging since it is not trivial to distinguish polyps from other problematic colonic tissues such as lesions, ulcerative colitis, and inflammatory bowel disease. Furthermore, many of the polyps presented in the dataset do not appear fully in the picture and are only visible in the corners, thus increasing the challenge. Further details regarding data splits of these datasets are provided in Table~\ref{tbl:dataset}.

\subsection{Models}
\label{sec:models}

To showcase the capabilities of the proposed method, we employ the most commonly used transformer-based computer vision architecture, Vision Transformer-Base/16 (ViT-B/16)~\cite{vit}. As suggested by~\cite{vit}, we use this model for images of size $224 \times 224$ with image tokens of size $16 \times 16$, resulting in a total of $196$ tokens. We modify the final linear layer of the model to accommodate the two-class classification problem tackled in this work.

We use two ViT-B/16 models where the first one is randomly initialized and trained from scratch while the second one is pretrained in a supervised fashion. Apart from those two models, in order to capture various properties of self-supervised learning methods that have seen increased use in medical imaging problems in recent years, we employ two additional models pretrained using (1) DINO~\cite{dino}, a discriminative SSL framework, and (2) MAE~\cite{mae}, which relies on a generative approach. The pretraining for the aforementioned models is performed on the ImageNet dataset~\cite{ILSVRC15:rus}, a large-scale dataset containing natural images.

\begin{figure}[t]
\centering
\begin{tikzpicture}
\centering

\def\sety1{0}
\node[inner sep=0pt] (a) at (0, \sety1)
{\includegraphics[width=1.6cm]{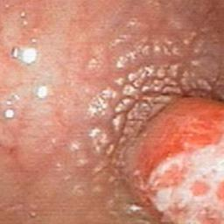}};
\node[align=center] at (0, \sety1 - 1.1) {\scriptsize $i{=}0$};
\node[align=center] at (0, \sety1 - 1.45) {\scriptsize $y'_c{=}0.99$};

\node[inner sep=0pt] (b) at (1.7, \sety1)
{\includegraphics[width=1.6cm]{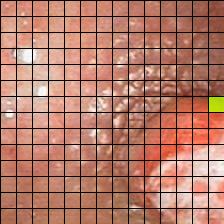}};
\node[align=center] at (1.7, \sety1 - 1.1) {\scriptsize $i{=}1, t{=}98$};
\node[align=center] at (1.7, \sety1 - 1.45) {\scriptsize $y'_c{=}0.99$};

\node[inner sep=0pt] (c) at (3.4, \sety1)
{\includegraphics[width=1.6cm]{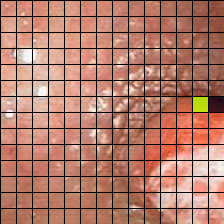}};
\node[align=center] at (3.4, \sety1 - 1.1) {\scriptsize $i{=}2, t{=}97$};
\node[align=center] at (3.4, \sety1 - 1.45) {\scriptsize $y'_c{=}0.99$};

\node[inner sep=0pt] (c) at (5.1, \sety1)
{\includegraphics[width=1.6cm]{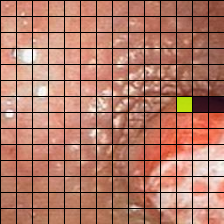}};
\node[align=center] at (5.1, \sety1 - 1.1) {\scriptsize $i{=}3, t{=}96$};
\node[align=center] at (5.1, \sety1 - 1.45) {\scriptsize $y'_c{=}0.99$};

\node[inner sep=0pt] (c) at (6.8, \sety1)
{\includegraphics[width=1.6cm]{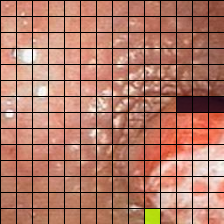}};
\node[align=center] at (6.8, \sety1 - 1.1) {\scriptsize $i{=}4, t{=}192$};
\node[align=center] at (6.8, \sety1 - 1.45) {\scriptsize $y'_c{=}0.99$};

\node[inner sep=0pt] (c) at (8.5, \sety1)
{\includegraphics[width=1.6cm]{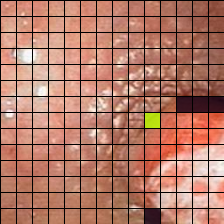}};
\node[align=center] at (8.5, \sety1 - 1.1) {\scriptsize $i{=}5, t{=}108$};
\node[align=center] at (8.5, \sety1 - 1.45) {\scriptsize $y'_c{=}0.99$};

\node[inner sep=0pt] (c) at (10.2, \sety1)
{\includegraphics[width=1.6cm]{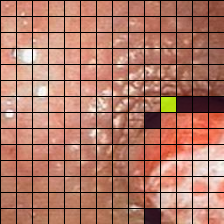}};
\node[align=center] at (10.2, \sety1 - 1.1) {\scriptsize $i{=}6, t{=}95$};
\node[align=center] at (10.2, \sety1 - 1.45) {\scriptsize $y'_c{=}0.99$};

\def\sety2{-2.5}
\node[inner sep=0pt] (a) at (0, \sety2)
{\includegraphics[width=1.6cm]{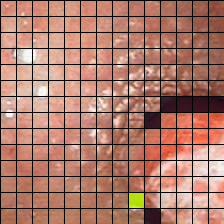}};
\node[align=center] at (0, \sety2 - 1.1) {\scriptsize $i{=}7, t{=}177$};
\node[align=center] at (0, \sety2 - 1.45) {\scriptsize $y'_c{=}0.99$};

\node[inner sep=0pt] (b) at (1.7, \sety2)
{\includegraphics[width=1.6cm]{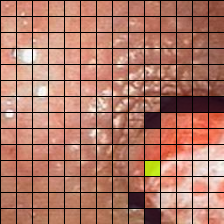}};
\node[align=center] at (1.7, \sety2 - 1.1) {\scriptsize $i{=}8, t{=}150$};
\node[align=center] at (1.7, \sety2 - 1.45) {\scriptsize $y'_c{=}0.99$};

\node[inner sep=0pt] (c) at (3.4, \sety2)
{\includegraphics[width=1.6cm]{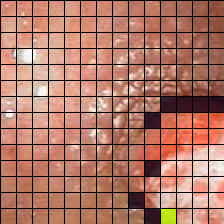}};
\node[align=center] at (3.4, \sety2 - 1.1) {\scriptsize $i{=}9, t{=}193$};
\node[align=center] at (3.4, \sety2 - 1.45) {\scriptsize $y'_c{=}0.98$};

\node[inner sep=0pt] (c) at (5.1, \sety2)
{\includegraphics[width=1.6cm]{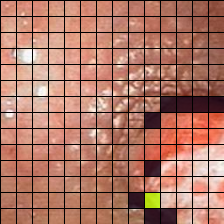}};
\node[align=center] at (5.1, \sety2 - 1.1) {\scriptsize $i{=}10, t{=}178$};
\node[align=center] at (5.1, \sety2 - 1.45) {\scriptsize $y'_c{=}0.97$};

\node[inner sep=0pt] (c) at (6.8, \sety2)
{\includegraphics[width=1.6cm]{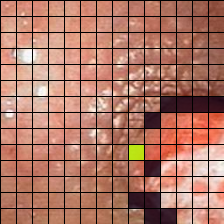}};
\node[align=center] at (6.8, \sety2 - 1.1) {\scriptsize $i{=}11, t{=}135$};
\node[align=center] at (6.8, \sety2 - 1.45) {\scriptsize $y'_c{=}0.96$};

\node[inner sep=0pt] (c) at (8.5, \sety2)
{\includegraphics[width=1.6cm]{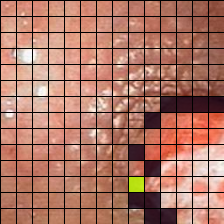}};
\node[align=center] at (8.5, \sety2 - 1.1) {\scriptsize $i{=}12, t{=}163$};
\node[align=center] at (8.5, \sety2 - 1.45) {\scriptsize $y'_c{=}0.94$};

\node[inner sep=0pt] (c) at (10.2, \sety2)
{\includegraphics[width=1.6cm]{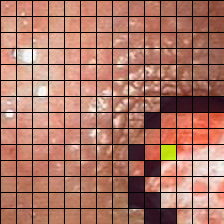}};
\node[align=center] at (10.2, \sety2 - 1.1) {\scriptsize $i{=}13, t{=}137$};
\node[align=center] at (10.2, \sety2 - 1.45) {\scriptsize $y'_c{=}0.91$};

\def\sety3{-5}
\node[inner sep=0pt] (a) at (0, \sety3)
{\includegraphics[width=1.6cm]{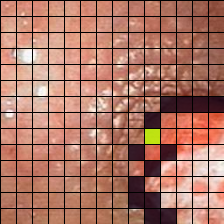}};
\node[align=center] at (0, \sety3 - 1.1) {\scriptsize $i{=}14, t{=}122$};
\node[align=center] at (0, \sety3 - 1.45) {\scriptsize $y'_c{=}0.87$};

\node[inner sep=0pt] (b) at (1.7, \sety3)
{\includegraphics[width=1.6cm]{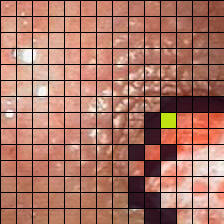}};
\node[align=center] at (1.7, \sety3 - 1.1) {\scriptsize $i{=}15, t{=}109$};
\node[align=center] at (1.7, \sety3 - 1.45) {\scriptsize $y'_c{=}0.81$};

\node[inner sep=0pt] (c) at (3.4, \sety3)
{\includegraphics[width=1.6cm]{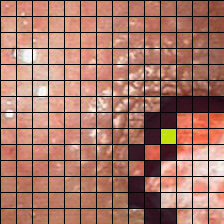}};
\node[align=center] at (3.4, \sety3 - 1.1) {\scriptsize $i{=}16, t{=}123$};
\node[align=center] at (3.4, \sety3 - 1.45) {\scriptsize $y'_c{=}0.76$};

\node[inner sep=0pt] (c) at (5.1, \sety3)
{\includegraphics[width=1.6cm]{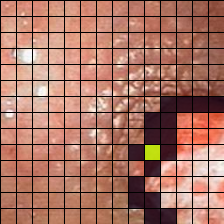}};
\node[align=center] at (5.1, \sety3 - 1.1) {\scriptsize $i{=}17, t{=}136$};
\node[align=center] at (5.1, \sety3 - 1.45) {\scriptsize $y'_c{=}0.66$};

\node[inner sep=0pt] (c) at (6.8, \sety3)
{\includegraphics[width=1.6cm]{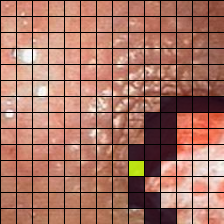}};
\node[align=center] at (6.8, \sety3 - 1.1) {\scriptsize $i{=}18, t{=}149$};
\node[align=center] at (6.8, \sety3 - 1.45) {\scriptsize $y'_c{=}0.57$};

\node[inner sep=0pt] (c) at (8.5, \sety3)
{\includegraphics[width=1.6cm]{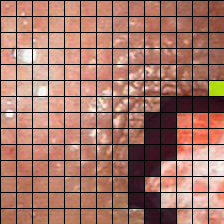}};
\node[align=center] at (8.5, \sety3 - 1.1) {\scriptsize $i{=}19, t{=}84$};
\node[align=center] at (8.5, \sety3 - 1.45) {\scriptsize $y'_c{=}0.50$};

\node[inner sep=0pt] (c) at (10.2, \sety3)
{\includegraphics[width=1.6cm]{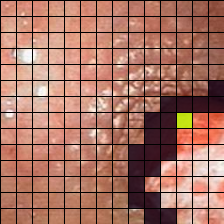}};
\node[align=center] at (10.2, \sety3 - 1.1) {\scriptsize $i{=}20, t{=}110$};
\node[align=center] at (10.2, \sety3 - 1.45) {\scriptsize\underline{$y'_c{=}0.44$}};

\end{tikzpicture}
\caption{An example of procedure of Token Insight where critical tokens ($t$) that contribute to the prediction made by the model are iteratively discovered based on their impact on prediction. Identified tokens are discarded after each iteration ($i$) until the prediction changes from positive to negative. For clarity, we provide the prediction confidence ($y'_c$) below each each image where this result is obtained using the image with discarded tokens (black patches). The token that is removed at each iteration is highlighted in green.}
\label{fig:token_insight_iterative}
\end{figure}
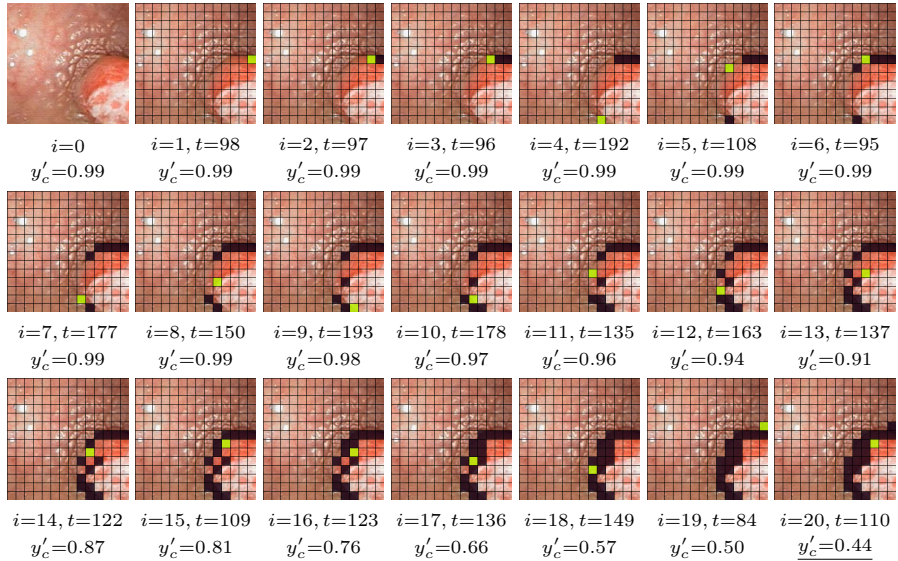

\subsection{Token Insight}
\label{sec:token_insight}

Given an image $\texttt{X} \in \mathbb{R}^{D\times W\times H}$ and its categorical association $\bm{y} \in \mathbb{R}^M$, sampled from a dataset $(\texttt{X}, \bm{y}) \sim \mathcal{D}$, where $y_c = 1$ and $y_m = 0$ for all $m \in \{0,\ldots, M\} \backslash {c}$, let $g_\theta(\cdot)$ represent a vision transformer with parameters $\theta$ that maps an image to a set of prediction likelihoods, denoted as $g(\theta, \mathtt{X}) = \bm{y}'$, where $\sum_{i=1}^{M} y_i' = 1$. If $\arg\max (\bm{y}') = c$, then the classification is considered correct.

When utilized as an input for a vision transformer, the image $\texttt{X}$ undergoes a process of patchification (also called tokenization), transforming it into a set of tokens as shown in Figure~\ref{fig:transformer_fig}. In this study, we adopt ViT-B/16 which employs $16\times 16$ patches. Given that the input images have a resolution of $224 \times 224$, the input $\texttt{X}$ is patchified into $196$ tokens denoted as $\mathtt{X} = [\bm{x}_1, \bm{x}_2, \ldots, \bm{x}_{196}]$. As shown in Figure~\ref{fig:transformer_fig}, in the setup of ViT, a special token called \texttt{[cls]} token is prepended on $\mathtt{X}$, thus increasing the number of tokens by one. This token is then used for the purpose of making classification using a linear layer after the final transformer encoder layer.

Token Insight is designed to progressively identify the most critical token contributing to the model's prediction. To achieve this, at each step, it removes one token at a time, measures the change in prediction, and identifies the token causing the largest drop in prediction confidence for the correct class. Subsequently, after examining each token, the one resulting in the highest confidence change when removed is permanently discarded. This process is then repeated to discover remaining critical tokens until the prediction ultimately changes. In the case of polyp detection problem, Token Insight finds the tokens that lead to the prediction of the \say{Polyp} class and the token discarding operation is carried out repeatedly until the prediction changes from \say{Polyp-positive} to \say{Polyp-negative} for the images considered. Token Insight finds its origins in the works of~\cite{madsen2023faithfulness,IA_RED}. However, unlike those approaches which stop after a single iteration, we continue to search for critical tokens greedily and discard them until the prediction is eventually changed, thus identifying all of the critical tokens that contribute to the prediction.

Formally, we can represent Token Insight as follows: given an input image represented as $\mathtt{X} = [\bm{x}_k]_{k \in \{1,\ldots,196\}}$, denote by $\mathtt{X}^{(t_1, \ldots, t_i)}$ the image where the tokens $\{t_1, \ldots, t_i\}$ have been removed. At the next iteration, $i + 1$, the most critical token is the token $t_{i + 1}$ that gives rise to the largest drop in prediction confidence:
\[
g(\theta, \mathtt{X}^{(t_1, \ldots, t_i, t_{i+1})})_c \le 
g(\theta, \mathtt{X}^{(t_1, \ldots, t_i, s)})_c, \quad \text{for all $s \in \{1,\ldots,196\} \backslash \{t_1, \ldots, t_i\}$}.
\]

The iterative search described above stops when $\arg \max (g(\theta, \mathtt{X}^{(t_1, \ldots, t_i)})_c) \neq c$. Note that Token Insight does not alter or remove the \texttt{[cls]} token during the process described above and, in order to generate the prediction confidences \texttt{[cls]} token is used according to the description of~\cite{vit} throughout the process.

An illustration of the token discarding operation of Token Insight is provided in Figure~\ref{fig:token_insight_iterative} where the example image is initially predicted with $0.99$ confidence as having a polyp. Using Token Insight, critical tokens are identified and discarded iteratively until the prediction eventually drops down to $0.44$ for the polyp-positive class, thus identifying tokens that contribute to the prediction made by the model.

\textbf{Relation to occlusion-based methods}\,\textendash\,
Note that, on the surface, Token Insight appears to be similar to the occlusion-based methods which mask certain regions of the input to measure the change in the prediction~\cite{zeiler2014visualizing}. The primary difference between  Token Insight and these methods is that during the token discarding process the model is not affected by the potential spurious signals such as missingness bias introduced by the masking operation (i.e., the bias that is introduced by the mask)~\cite{jain2022missingness}. Since the tokens are completely removed, the model prediction is not influenced by the removed tokens. Unfortunately, since the token removal operation is only available for transformer-based models, Token Insight is not usable for CNNs and other non-transformer DNN architectures.

\section{Experimental Results}
\subsection{Training}

As described in Section~\ref{sec:models} and Section~\ref{sec:dataset}, we employ four ViT-B/16 models, three of which are pretrained on the ImageNet dataset, and train them on the combined CP-Child dataset. To discover the most appropriate model with the highest performance, we perform extensive training efforts  and in what follows, we detail the training routine that leads to the identification of the best model.

For pretrained models (supervised, DINO, MAE), training is performed for $25$ epochs using Stochastic Gradient Descent (SGD) with an initial learning rate of $10^{-3}$ and momentum of $0.05$. The model trained from scratch undergoes a longer training period ($50$ epochs) to compensate for the lack of pretraining and uses an increased learning rate of $0.035$, weight decay of $10^{-5}$, and momentum of $0.075$. All models employ a batch size of $32$ and use cosine annealing, which reduces the learning rate after each epoch using the method described in~\cite{simsiam}.

In the work of~\cite{wang2020improved}, pretrained ResNets are reported to attain an accuracy of approximately $99\%$ on the CP-Child dataset. With pretrained ViTs, we replicate these results, achieving comparable performance. The model trained from scratch exhibits slightly reduced performance at $95.3\%$ due to the absence of pretraining. Nonetheless, this model still delivers commendable results on this dataset, making these models suitable for a study on predictive reasoning.

\begin{figure}[t!]
    \centering
    \begin{subfigure}{0.4\textwidth}
        \centering
        \includegraphics[width=0.75\textwidth]{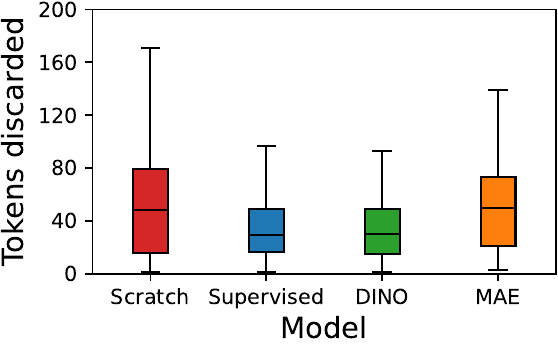}
        \caption{}
        \label{fig:tokens_needed}
        \includegraphics[width=0.75\textwidth]{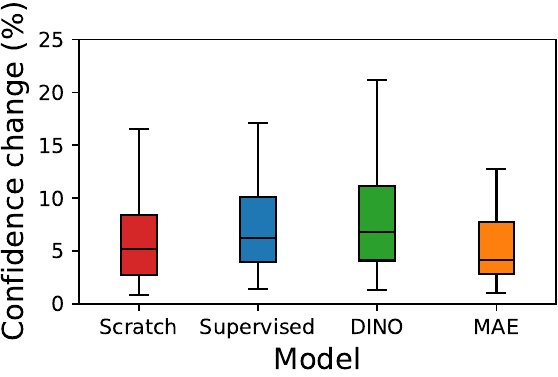}
        \caption{}
        \label{fig:max_conf_change}
    \end{subfigure}
    \begin{subfigure}{0.58\textwidth}
        \centering
        \includegraphics[width=0.45\textwidth]{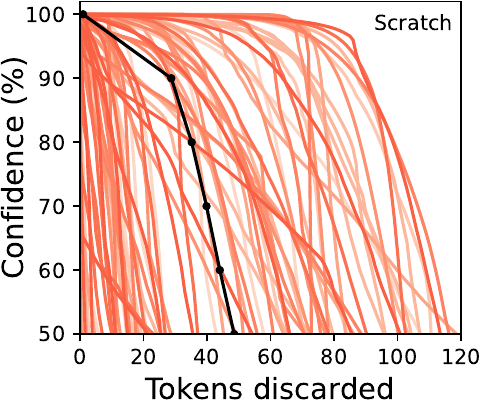}
        \includegraphics[width=0.45\textwidth]{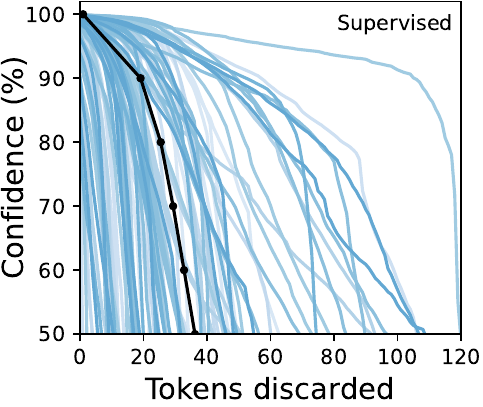}
        \includegraphics[width=0.45\textwidth]{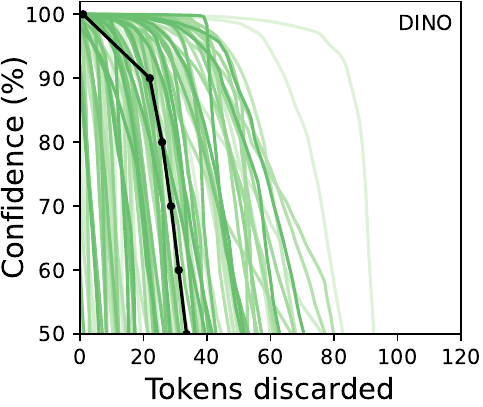}
        \includegraphics[width=0.45\textwidth]{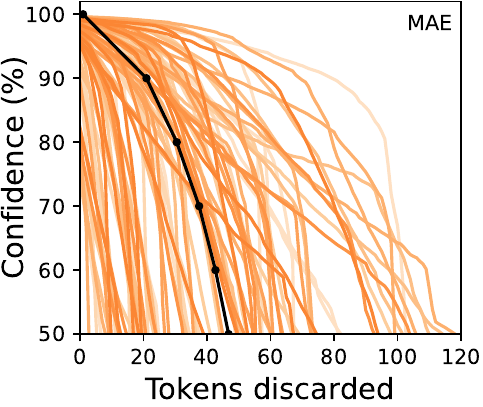}
        \caption{}
        \label{fig:lineplots}
    \end{subfigure}
    \caption{(a) Number of tokens discarded to change predictions from polyp-positive to polyp-negative for CP-Child dataset. (b) Maximum confidence change measured with a single token during the process of (a). Outliers are omitted to enhance visual clarity.
    (c) Change in polyp-positive confidence based on token removal for all images in the CP-Child test dataset with  black lines indicating average number of tokens discarded measured across all images.}
    \label{fig:exp_results}
\end{figure}

\subsection{Token Insight}

In order to investigate various properties of the proposed approach as well as models pretrained in different ways, we apply Token Insight to the images containing polyps in the CP-Child-B dataset.

\textbf{Number of tokens discarded}\,\textendash\,
For all four models, we investigate the number of tokens discarded via Token Insight to examine the model's reliance on tokens for making a polyp-positive prediction. Doing so, in Figure~\ref{fig:tokens_needed}, we present the number of tokens discarded to change a polyp-positive prediction into a polyp-negative one. As it can be seen, both MAE and the model trained from scratch discard more tokens before the prediction changes, while models pretrained with supervised learning and DINO discard fewer, and hence rely on more tokens to make a prediction. This implies that treating all transformer-based medical AI models the same would be a mistake since the number of tokens used for confirming a prediction may vary significantly depending on how the model is trained.

\textbf{The most impactful token}\,\textendash\,Expanding our investigation, in Figure~\ref{fig:max_conf_change}, we illustrate the influence of the most impactful tokens across the dataset, as measured by their impact on the prediction confidence when removed. Complementing our previous insights, we observe that individual tokens in the models pretrained with supervised learning, as well as DINO, have more influence on the prediction compared to MAE and the model trained from scratch. Once again, this observation reveals that it is possible to have two models sharing the same architecture, but where one heavily relies on fewer tokens, whereas the other focuses on a broader area. Consequently, depending on the medical imaging task at hand, practitioners may prefer one type of model over another.

Combining the two experiments discussed above, we present Figure~\ref{fig:lineplots} which shows confidence drops per removed token over the number of tokens discarded for each model. Graphs presented in Figure~\ref{fig:lineplots} can be seen as a summarizing views of Figure~\ref{fig:tokens_needed} and Figure~\ref{fig:max_conf_change}, where the goal is to reveal the different behaviors of various models. As can be seen, all models exhibit an initial stage in which confidence drops only moderately as the initial few tokens are removed, followed by a steep drop in confidence. MAE and scratch see a relative robustness in the decrease of prediction confidence as a function of the number of tokens removed, compared to DINO and supervised pretrained models.

\begin{figure}[t]
\centering
\begin{tikzpicture}
\centering

\node[align=center] at (-1.05, 0) {\rotatebox{90}{\scriptsize Image}};
\node[inner sep=0pt] (a) at (0, 0)
{\includegraphics[width=1.6cm]{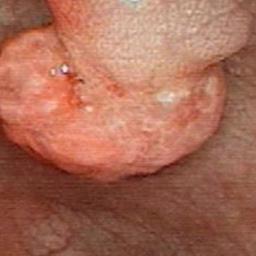}};
\node[inner sep=0pt] (a) at (1.7, 0)
{\includegraphics[width=1.6cm]{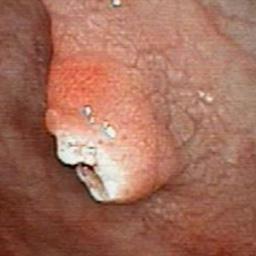}};
\node[inner sep=0pt] (a) at (3.4, 0)
{\includegraphics[width=1.6cm]{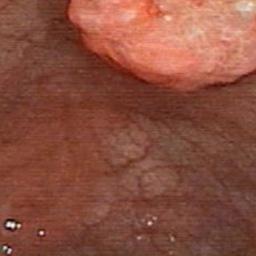}};
\node[inner sep=0pt] (a) at (5.1, 0)
{\includegraphics[width=1.6cm]{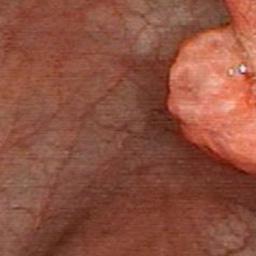}};
\node[inner sep=0pt] (a) at (6.8, 0)
{\includegraphics[width=1.6cm]{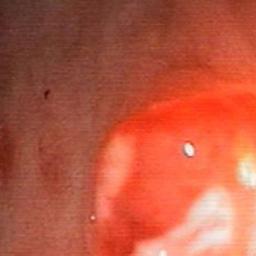}};
\node[inner sep=0pt] (a) at (8.5, 0)
{\includegraphics[width=1.6cm]{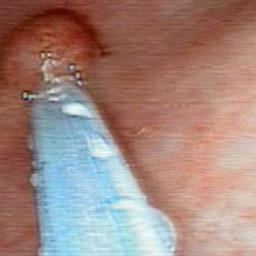}};
\node[inner sep=0pt] (a) at (10.2, 0)
{\includegraphics[width=1.6cm]{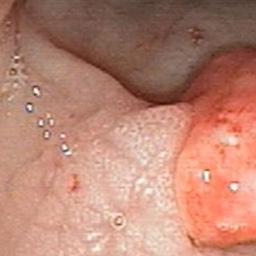}};

\node[align=center] at (-1.05, -1.7) {\rotatebox{90}{\scriptsize Scratch}};
\node[inner sep=0pt] (a) at (0, -1.7)
{\includegraphics[width=1.6cm]{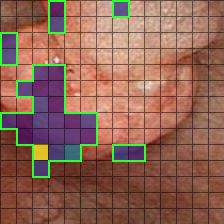}};
\node[inner sep=0pt] (a) at (1.7, -1.7)
{\includegraphics[width=1.6cm]{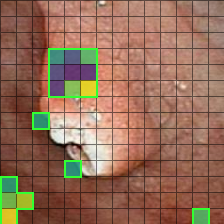}};
\node[inner sep=0pt] (a) at (3.4, -1.7)
{\includegraphics[width=1.6cm]{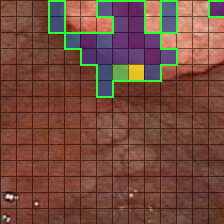}};
\node[inner sep=0pt] (a) at (5.1, -1.7)
{\includegraphics[width=1.6cm]{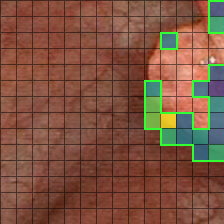}};
\node[inner sep=0pt] (a) at (6.8, -1.7)
{\includegraphics[width=1.6cm]{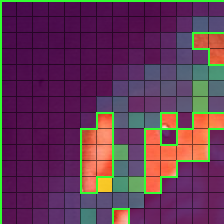}};
\node[inner sep=0pt] (a) at (8.5, -1.7)
{\includegraphics[width=1.6cm]{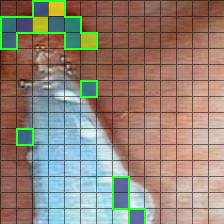}};
\node[inner sep=0pt] (a) at (10.2, -1.7)
{\includegraphics[width=1.6cm]{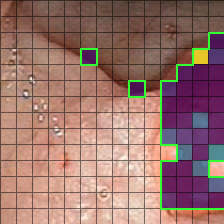}};

\node[align=center] at (-1.05, -3.4) {\rotatebox{90}{\scriptsize Supervised}};
\node[inner sep=0pt] (a) at (0, -3.4)
{\includegraphics[width=1.6cm]{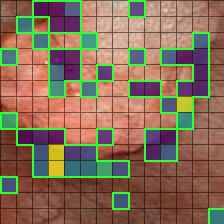}};
\node[inner sep=0pt] (a) at (1.7, -3.4)
{\includegraphics[width=1.6cm]{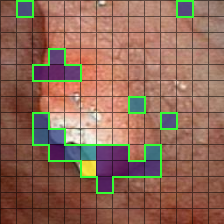}};
\node[inner sep=0pt] (a) at (3.4, -3.4)
{\includegraphics[width=1.6cm]{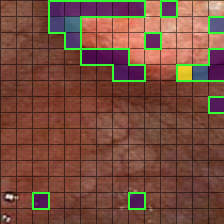}};
\node[inner sep=0pt] (a) at (5.1, -3.4)
{\includegraphics[width=1.6cm]{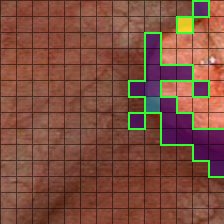}};
\node[inner sep=0pt] (a) at (6.8, -3.4)
{\includegraphics[width=1.6cm]{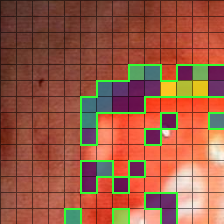}};
\node[inner sep=0pt] (a) at (8.5, -3.4)
{\includegraphics[width=1.6cm]{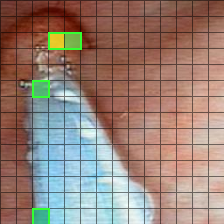}};
\node[inner sep=0pt] (a) at (10.2, -3.4)
{\includegraphics[width=1.6cm]{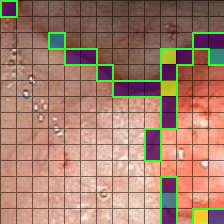}};

\node[align=center] at (-1.05, -5.1) {\rotatebox{90}{\scriptsize DINO}};
\node[inner sep=0pt] (a) at (0, -5.1)
{\includegraphics[width=1.6cm]{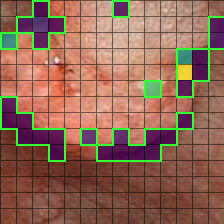}};
\node[inner sep=0pt] (a) at (1.7, -5.1)
{\includegraphics[width=1.6cm]{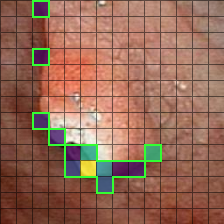}};
\node[inner sep=0pt] (a) at (3.4, -5.1)
{\includegraphics[width=1.6cm]{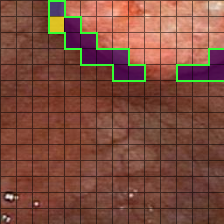}};
\node[inner sep=0pt] (a) at (5.1, -5.1)
{\includegraphics[width=1.6cm]{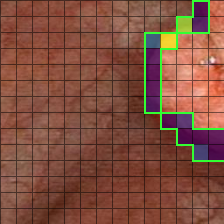}};
\node[inner sep=0pt] (a) at (6.8, -5.1)
{\includegraphics[width=1.6cm]{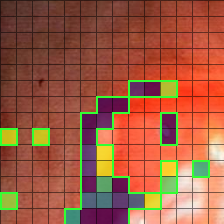}};
\node[inner sep=0pt] (a) at (8.5, -5.1)
{\includegraphics[width=1.6cm]{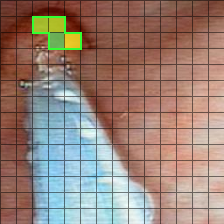}};
\node[inner sep=0pt] (a) at (10.2, -5.1)
{\includegraphics[width=1.6cm]{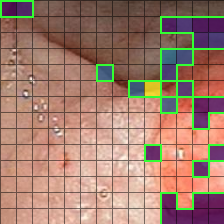}};

\node[align=center] at (-1.05, -6.8) {\rotatebox{90}{\scriptsize MAE}};
\node[inner sep=0pt] (a) at (0, -6.8)
{\includegraphics[width=1.6cm]{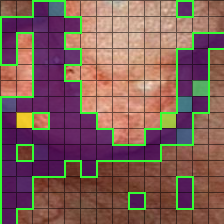}};
\node[inner sep=0pt] (a) at (1.7, -6.8)
{\includegraphics[width=1.6cm]{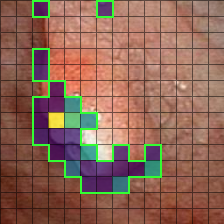}};
\node[inner sep=0pt] (a) at (3.4, -6.8)
{\includegraphics[width=1.6cm]{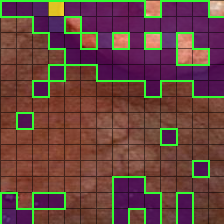}};
\node[inner sep=0pt] (a) at (5.1, -6.8)
{\includegraphics[width=1.6cm]{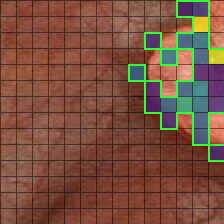}};
\node[inner sep=0pt] (a) at (6.8, -6.8)
{\includegraphics[width=1.6cm]{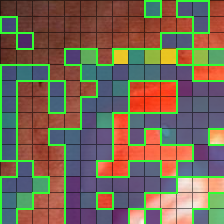}};
\node[inner sep=0pt] (a) at (8.5, -6.8)
{\includegraphics[width=1.6cm]{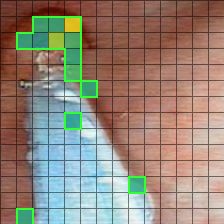}};
\node[inner sep=0pt] (a) at (10.2, -6.8)
{\includegraphics[width=1.6cm]{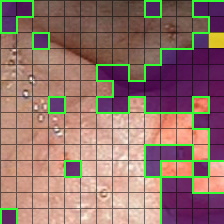}};

\end{tikzpicture}
\caption{Illustrations of Token Insight maps highlighting the most impactful tokens for a polyp-positive prediction based on models pretrained with various methods.}
\label{fig:token_insight_results}
\end{figure}
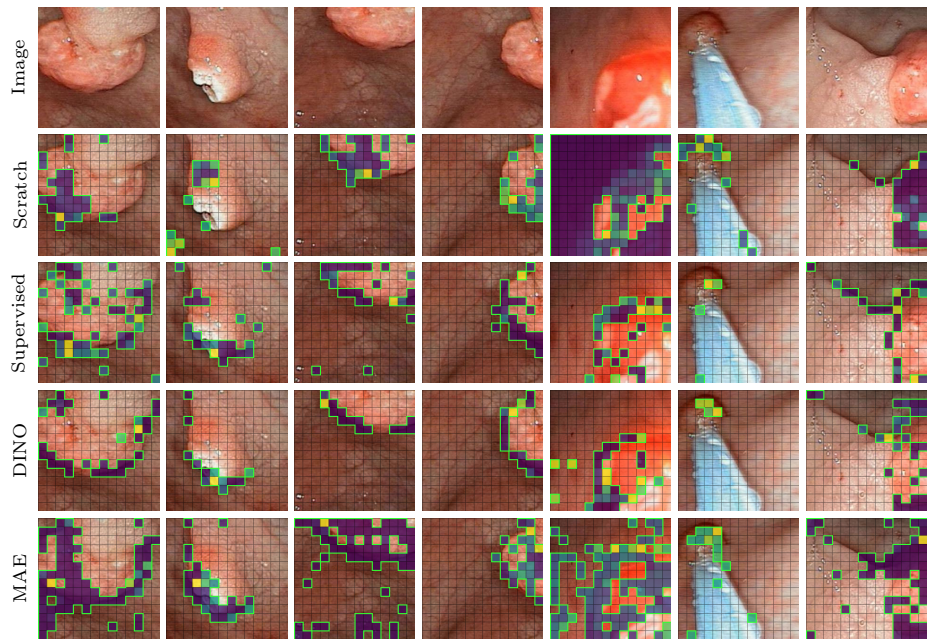

\textbf{Identifying shortcut learning}\,\textendash\,In Figure~\ref{fig:token_insight_results}, we present several examples showcasing the application of Token Insight on images containing polyps. Note that, while Token Insight accurately identifies regions containing polyps, thereby indicating that the model learns the correct signals, there are instances (particularly for MAE and Scratch) where certain tokens are identified as important but are not medically relevant to the prediction. This suggests that the model under consideration is indeed making predictions based on signals that are not intended in the dataset, also identified by~\cite{sun2023right}. As demonstrated, the primary use-case of the proposed approach is to uncover such cases and to discover predictions based on erroneous signals, thereby enhancing the trustworthiness of AI-based medical imaging.

\section{Conclusion}

In this work, we introduced Token Insight, a novel computational method designed to identify critical tokens that contribute to the prediction for transformer-based medical imaging models, thus enabling its usage as a method for identifying spurious correlations as well as erroneous signals that lead to shortcut learning. Unlike many of its predecessors, this method does not necessitate any changes in the model or require any additional extra modules. Furthermore, the results obtained by the proposed approach rely entirely on the change in prediction confidence of the model, making it a reliable indicator for identifying the undesired cases mentioned above.

We foresee two possible directions for future work. The current method identifies a set of tokens that influence the prediction by greedily removing the highest impact token at every iteration, and we have offered computational evidence that the set thus obtained forms a good approximation to the smallest set of highly influential tokens. For the future, we plan to investigate theoretical bounds for how well our algorithm manages to capture the smallest set of tokens that influence the prediction. A second line of work concerns the computational cost of our method, which scales as $O(N^2)$ in the number of tokens. Given the considerable computational cost involved in a forward pass of a transformer-based model, it would be of interest to re-use previously acquired information about the impact of a token to avoid making a forward pass for each candidate token.

\section{Acknowledgements}
This work was supported by a grant from the Special Research Fund (BOF) of Ghent University (BOF/STA/202109/039).

\bibliographystyle{splncs04}
\bibliography{new_main}

\end{document}